\documentclass[letterpaper]{article} 
\usepackage{aaai23}  
\usepackage{times}  
\usepackage{helvet}  
\usepackage{courier}  
\usepackage[hyphens]{url}  
\usepackage{graphicx} 
\usepackage{natbib}  
\usepackage{caption} 
\usepackage{algorithm}
\usepackage{newfloat}
\usepackage{listings}

\usepackage{algpseudocode}
\usepackage{amssymb}

\usepackage{enumitem}

\DeclareCaptionStyle{ruled}{labelfont=normalfont,labelsep=colon,strut=off} 
\floatstyle{ruled}
\newfloat{listing}{tb}{lst}{}
\floatname{listing}{Listing}
\usepackage{color}

\usepackage{graphicx}
\usepackage{amsmath}
\usepackage{amssymb}
\usepackage{booktabs}

\usepackage{makecell,tabularx}
\usepackage{multirow}

\usepackage{tikz}
\usepackage{comment}
\usepackage{colortbl}
\definecolor{mygray}{RGB}{230,230,250}
\definecolor{mygray2}{RGB}{176,196,222}
\definecolor{dblue}{RGB}{0, 112, 192}
\definecolor{dred}{RGB}{192, 0, 0}
\definecolor{Gray}{gray}{0.9}
\definecolor{tblue}{HTML}{174992}
\definecolor{barca-blue}{RGB}{0, 76, 153}
\definecolor{barca-red}{RGB}{167, 0, 66}
\definecolor{lblue}{RGB}{13, 152, 255}
\definecolor{lred}{RGB}{255, 108, 108}
\newcommand{\etal}{\textit{et al}.}
\usepackage{arydshln}

\title{Parameter-efficient Model Adaptation for Vision Transformers}
\author{
    Xuehai He\textsuperscript{\rm 1},
    Chunyuan Li\textsuperscript{\rm 2},
    Pengchuan Zhang\textsuperscript{\rm 2},
    Jianwei Yang\textsuperscript{\rm 2},
    Xin Eric Wang\textsuperscript{\rm 1}
}
\affiliations{
    \textsuperscript{\rm 1} UC Santa Cruz,
    \textsuperscript{\rm 2}Microsoft Research at Redmond\\
  \texttt{\{xhe89,xwang366\}@ucsc.edu, \{chunyl,penzhan,jianwyan\}@microsoft.com} 
}
\usepackage{bibentry}
\begin{document}
\maketitle
\begin{abstract}
In computer vision, it has achieved great transfer learning performance via adapting large-scale pretrained vision models (e.g., vision transformers) to downstream tasks. Common approaches for model adaptation either update all model parameters or leverage linear probes. In this paper, we aim to study parameter-efficient model adaptation strategies for vision transformers on the image classification task. We formulate efficient model adaptation as a subspace training problem and perform a comprehensive benchmarking over different efficient adaptation methods. We conduct an empirical study on each efficient model adaptation method focusing on its performance alongside parameter cost. Furthermore, we propose a parameter-efficient model adaptation framework, which first selects submodules by measuring local intrinsic dimensions and then projects them into subspace for further decomposition via a novel Kronecker Adaptation (KAdaptation) method. We analyze and compare our method with a diverse set of baseline model adaptation methods (including state-of-the-art methods for pretrained language models). Our method performs the best in terms of the tradeoff between accuracy and parameter efficiency across 20 image classification datasets under the few-shot setting and 7  image classification datasets under the full-shot setting. 
\end{abstract}

\section{Introduction}
In the last few years, large-scale vision models and language models pretrained on web-scale data have seen a great surge of interest with promising performance~\cite{gpt2,bert,yang2019xlnet,liu2019roberta}. Meanwhile, aided by the rapid gains in hardware, their sizes keep growing rapidly. Currently, vision transformers~\cite{vision_transformer} (ViTs) with billions of parameters such as {\em ViT-Large}~\cite{vision_transformer} have been released. 
It is expected that pretrained vision models with even larger orders of magnitude will emerge in the foreseeable future. 

\begin{figure}[t]
	\begin{center}
 	\includegraphics[width = 0.8\columnwidth]{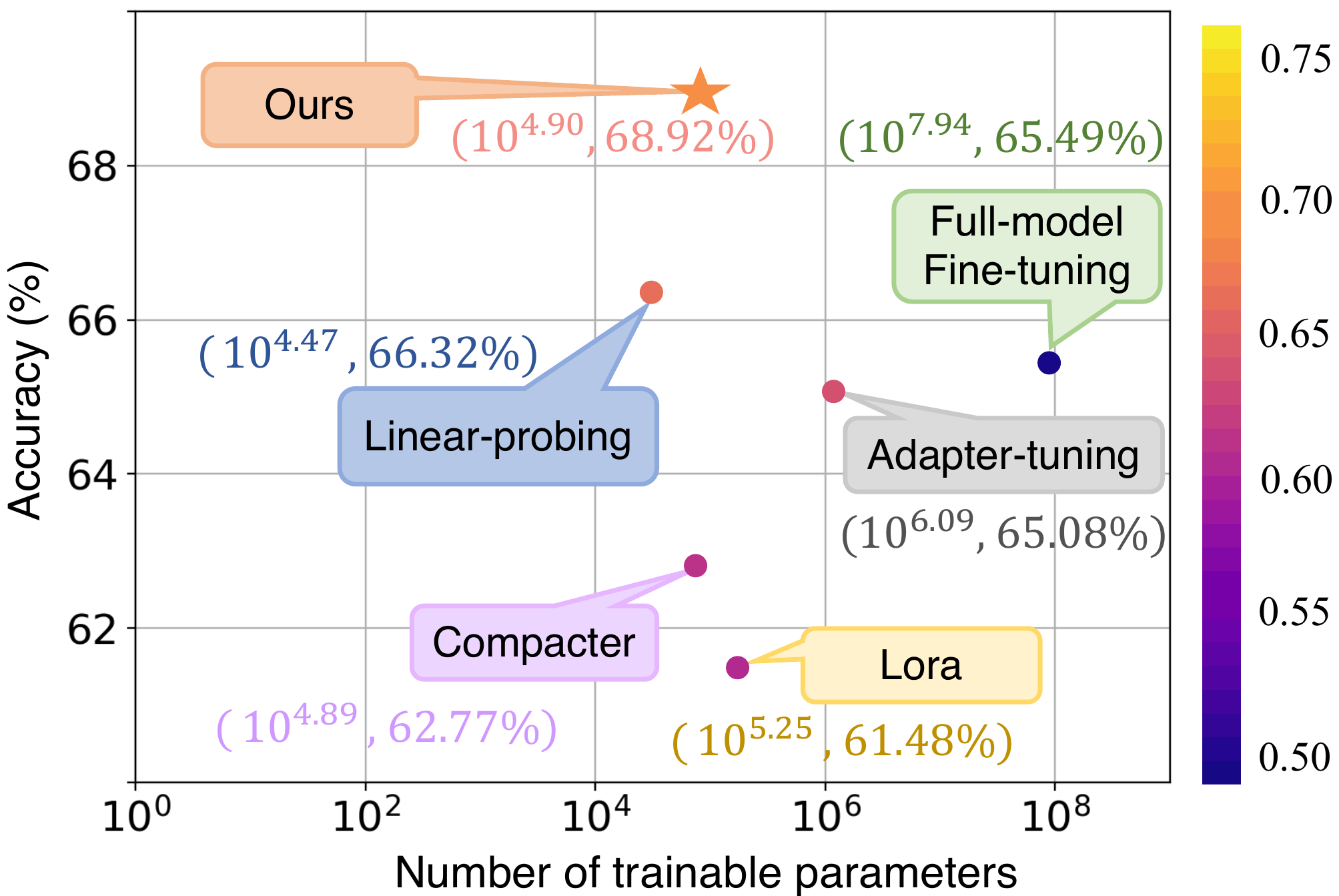}
 	\caption{The tradeoff between accuracy and parameter numbers of various model adaptation methods. The results are measured using the vision transformer
(ViT-B-224/32) via CLIP pretraining across the average of 20 image classification datasets. Our method places in the topleft corner and achieves the best tradeoff between accuracy and parameter efficiency. The color of points and numbers denote the performance-efficiency (PE) metric (higher is better). }\label{fig:trend}
 	\vspace{-0.2cm}
	\end{center}
 \end{figure}

These large-scale pretrained models are powerful when transferred to downstream vision tasks. However, deploying many independent instances of fine-tuned models can also cause substantial storage and deployment costs and hinder the applicability of large-scale ViTs to real-world problems. Motivated by this and the importance of parameter-efficient learning~\cite{houlsbyParameterEfficientTransferLearning2019,huLoRALowRankAdaptation2021,zakenBitFitSimpleParameterefficient2021,mahabadiCompacterEfficientLowRank2021,he2021towards}, we aim to study the parameter-efficient model adaptation strategy for vision transformers. Conventional wisdom for transfer learning in our computer vision community is fine-tuning all model parameters or leveraging linear probes. However, performing full-model fine-tuning of pretrained ViTs may incur both financial and environmental costs~\cite{patterson2021carbon}, requires a high computational budget, and becomes increasingly infeasible as the model size continuously grows. Another go-to strategy is performing linear-probing by stacking an additional trainable multi-layer perceptron (MLP) layer in the end. It is parameter-efficient yet suboptimal in terms of performance. 
Ideally, we hope to design model adaptation strategies that can achieve the best tradeoff between efficiency and effectiveness (see Figure~\ref{fig:trend}) --- optimizing adaptation parameter-efficiency while allowing for the model to maintain the effectiveness of transfer learning on downstream vision tasks, especially the image classification task.
 
To this end, \textbf{we ask} an essential question: {\em what are the general guidelines one should adopt while adapting large-scale pretrained vision models on the downstream image classification datasets?}  
This work aims to answer the question by building a benchmark for adapting ViTs and proposing a more parameter-efficient model adaptation method. 
We choose ViTs as the pretrained vision models, which are representative mainstream state-of-the-art (SOTA) models on a wide range of downstream vision tasks. 
Specifically, we experiment with two off-the-shelf pretrained ViTs in the remainder of this paper: the one via Contrastive Language-Image Pretraining (also known as CLIP)~\cite{clip}, and the one via supervised pretraining (we refer to as Supervised ViT)~\cite{Transformers}. In addition to Full-model Fine-tuning and linear-probing, we re-implement several SOTA efficient adaptation methods~\cite{houlsbyParameterEfficientTransferLearning2019,ruckleAdapterDropEfficiencyAdapters2021,huLoRALowRankAdaptation2021,zakenBitFitSimpleParameterefficient2021,liPrefixTuningOptimizingContinuous2021} (originally proposed for pretrained language models) on vision tasks, and design various new baseline methods for comparison. 

Aghajanyan~\etal~\shortcite{aghajanyanIntrinsicDimensionalityExplains2020}  show that pretrained language models have a low intrinsic dimension and can still learn efficiently despite a low-dimensional reparameterization. 
Motivated by this observation, we reformulate the task of efficient model adaptation as a subspace training problem.
Within this framework, we measure the \emph{local intrinsic dimension} of each module in ViTs, which reveals that the attention module dominates the training progress. 
Moreover, we introduce a novel parameter-efficient model adaptation framework named \emph{Kronecker Adaptation (KAdaptation)}, where during adaptation, pretrained weights are frozen, and only the updates to the weights receive gradients.  And the weight updates are decomposed to a set of Kronecker products, with the {\em slow weights}~\cite{fast_weights} shared across layers and {\em fast weights}~\cite{fast_weights} further decomposed into low-rank matrices product to improve parameter efficiency. 
We apply KAdaptation to attention weights, and it achieves the best average accuracy among efficient model adaptation methods while containing much less trainable parameters,  e.g., around \textbf{45\%} parameters of LoRA~\cite{huLoRALowRankAdaptation2021} and \textbf{0.09\%} of all the model parameters in CLIP under the few-shot setting. 

The contributions of this paper are summarized below:
\begin{itemize}
\item We build a benchmark\footnote{To facilitate future research, implementations of all the methods studied in this work are released at~\url{https://github.com/eric-ai-lab/PEViT}.} for parameter-efficient model adaptation of ViTs on the image classification task by introducing our new baseline methods and several state-of-the-art efficient model adaptation strategies inspired from the NLP community. To our best knowledge, this is the first empirical study of the efficient model adaptation of Transformers to date that considers pure vision tasks.
\item We formulate efficient model adaptation as a subspace training problem. To address it, we define the local intrinsic dimension, based on which we choose submodules --- attention modules and we employ the proposed KAdaptation method to decompose the weight updates of attention modules for trainable parameter deduction.
\item We experiment  on 20 datasets under the few-shot setting and 7 image classification datasets under the full-shot setting. The results demonstrate the effectiveness of our method, achieving the best tradeoff between accuracy and parameter efficiency, as shown in Figure~\ref{fig:trend}.
\end{itemize}

\section{Related Work}
\paragraph{Vision Transformer}
Fine-tuning large-scale pretrained ViTs has shown prominent performance for computer vision tasks, such as image classification~\cite{vision_transformer}, object detection~\cite{detection_transformer}, and etc. Recently, there are also other variants, including hierarchical ViTs with varying resolutions and spatial embeddings~\cite{liuSwinTransformerHierarchical2021,dong2021cswin} been proposed. Undoubtedly, the recent progress of large ViTs posts great demands for developing efficient model adaptation strategies.

\paragraph{Efficient Model Adaptation in NLP}
In the natural language processing domain, efficient model
adaptation techniques typically involve adding to or modifying a limited number of parameters of the model — limiting the dimension of the optimization problem can prevent catastrophic forgetting~\cite{catastrophic}. Exiting methods are mainly divided into two categories depending on whether new trainable parameters are introduced. 
Specifically, one is to train a subset of the model parameters, where the common approach is to use a linear probe on top of pretrained features~\cite{clip}. The other alternatives include new parameters in between the network~\cite{liPrefixTuningOptimizingContinuous2021,ruckleAdapterDropEfficiencyAdapters2021,houlsbyParameterEfficientTransferLearning2019,huLoRALowRankAdaptation2021,pfeifferAdapterFusionNonDestructiveTask2021, vladapter}.
Nevertheless, these methodologies normally have not been investigated in the computer vision scenario and it is furthermore uncertain if findings from NLP tasks (e.g., question answering~\cite{rajpurkar2016squad}, natural language understanding~\cite{wang2018glue}, etc.) can transfer to downstream vision applications. Spurred by those facts, we establish a benchmark to compare these methods and we further advocate our method which can gain a better tradeoff under both the full-shot and few-shot settings. 

\section{Efficient Model Adaptation with Subspace Training}

Given a large pretrained vision transformer $\mathcal{M}$ with size $\vert \mathcal{M} \vert$. Our goal is to develop a parameter-efficient model adaptation technique with trainable parameters $\theta$ of size $d \ll \vert \mathcal{M} \vert$, that can attain comparable performance with fine-tuning the whole model. Our ultimate goal is that one could
achieve satisfactory results in both efficacy and efficiency without the hassle of fine-tuning the full model.

\subsection{Subspace Training}
A typical neural network contains numerous dense layers that perform matrix multiplication. The weight matrices in these layers can be full-rank. When adapting to a specific task, however, Aghajanyan~\etal~\shortcite{aghajanyanIntrinsicDimensionalityExplains2020} show that the pretrained language models have a low \emph{intrinsic dimension} and can still learn efficiently despite a low-dimensional reparameterization.

Drawing inspiration from their observation and study, we hypothesize that the updates to weights of ViTs during each step in model adaptation also have a low intrinsic rank and develop our method accordingly. The intuition behind our method is to perform subspace training on weight updates. In the de-facto training paradigm of neural network models, the gradient is computed first, followed by gradient steps taken by the optimizer in the entire parameter space $D$. While in subspace training, we instead build a random $d$-dimensional parameter subspace from $ \mathcal{M}$, where generally $d \ll \vert \mathcal{M} \vert$, and optimize directly in this subspace.

In fact, most current parameter-efficient NLP model adaptation strategies perform subspace training. Given a large pretrained language model $\mathcal{M}$ with size $\vert \mathcal{M} \vert$, existing methods either select a submodule from $\mathcal{M}$ or inject an additional module to $\mathcal{M}$. For the parameter vector $\Theta\in \mathbb{R}^{D}$ from this module, they learn a projection $\mathcal{P}$ mapping $\Theta$ into a random $d$-dimensional subspace and perform training in that subspace to minimize computational cost. With this observation, we motivate our study on the efficient model adaptation problem in the principle of subspace training. We approach the problem by addressing two scientific questions:  \emph{how to choose these submodules} and \emph{how to make the subspace projection}.

\subsection{The Proposed Kronecker Adaptation} 
To answer the two fundamental questions of efficient model adaptation, \emph{how to choose these submodules} and \emph{how to make the subspace projection},  we propose a novel framework that consists of two corresponding strategies. First, we define the local intrinsic dimension and we choose submodules based on their measured local intrinsic dimensions. Second, we propose a Kronecker Adaptation method to perform the subspace projection on the selected submodules by exploiting parameterized hypercomplex multiplication layers (PHM)~\cite{phm}.  

\subsubsection{Local Intrinsic Dimension} 
Measuring the intrinsic dimension of an objective function was first proposed in Li~\etal~\shortcite{liMeasuringIntrinsicDimension2018}. Aghajanyan~\etal~\shortcite{aghajanyanIntrinsicDimensionalityExplains2020} extended it to analyze the quality of pretrained language models. They point out that analyzing model adaptation through the lens of intrinsic dimension offers empirical and theoretical intuitions. Both of them study the intrinsic dimension of the entire model. 

Unlike them, we propose to measure the intrinsic dimension of each individual submodule in ViT. We define the intrinsic dimension of the submodule as \emph{local intrinsic dimension}, to distinguish it from the intrinsic dimension of the whole model. The local intrinsic dimension is indicative of the contribution of each submodule during model adaptation and measuring it will tell us how many free parameters are required to approximate the optimization problem closely. The conventional standard method of measuring the intrinsic dimensionality of an objective~\cite{liMeasuringIntrinsicDimension2018} asks for performing grid search over different subspace dimensions $d$, training using standard SGD~\cite{sgd} over the subspace reparameterization, and selecting the smallest $d$ which can produce a satisfactory solution (e.g., 90\% of the full training metric). 
Likewise, we measure the local intrinsic dimension via finding the smallest $d$ for the measured submodule that can reach 90\% of the full accuracy.

To this end, we first follow the similar definition in Li~\etal~\shortcite{liMeasuringIntrinsicDimension2018} and define $\Theta$ in a subspace in the following way: \begin{equation}
\Theta=\Theta_{0}+P \theta,
\end{equation}
where $\Theta_{0}\in \mathbb{R}^{D}$ is the initial parameter vector of $\Theta$ when the training begins, $P\in \mathbb{R}^{D \times d}$ is the projection matrix generated by the Fastfood transform~\cite{le2014fastfood}, and $\theta \in \mathbb{R}^{d}$ is the parameter vector in the subspace. Subspace training proceeds by computing gradients with respect to $\theta$ and taking steps in that subspace. By performing experiments with gradually larger values of $d$, we can find the subspace dimension $d_t$ at which the performance of the model $\mathcal{M}$ reaches 90\% of the full accuracy. We refer to $d_t$ the \emph{local intrinsic dimension} of the measured submodule.

The module with the lowest local intrinsic dimension --- attention module is selected. We project them into subspace via our proposed KAdaptation method for the sake of efficient model adaptation.
KAdaptation fine-tunes attention weight matrices indirectly by optimizing decomposition matrices of the updates to attention weight matrices. To lower the parameter cost, the decomposition is computed as the sum of Kronecker products while the original matrices remain frozen.

\subsubsection{Kronecker Product}
The Kronecker product between matrix $\boldsymbol{A} \in \mathbb{R}^{m \times n}$ and $\boldsymbol{B} \in \mathbb{R}^{p \times q}$, denoted by $\boldsymbol{A} \otimes \boldsymbol{B} \in \mathbb{R}^{m p \times n q}$, is mathematically written in the following form:
\begin{equation}
\boldsymbol{A} \otimes \boldsymbol{B}=\left(\begin{array}{ccc}
a_{11} \boldsymbol{B} & \cdots & a_{1 n} \boldsymbol{B} \\
\vdots & \ddots & \vdots \\
a_{m 1} \boldsymbol{B} & \cdots & a_{m n} \boldsymbol{B},
\end{array}\right)
\end{equation}
where $a_{i j}$ shows the element in the $i$-{th} row and $j$-{th} column of $\boldsymbol{A}$.

\begin{figure}[t]
	\begin{center}
 	\includegraphics[width = \columnwidth]{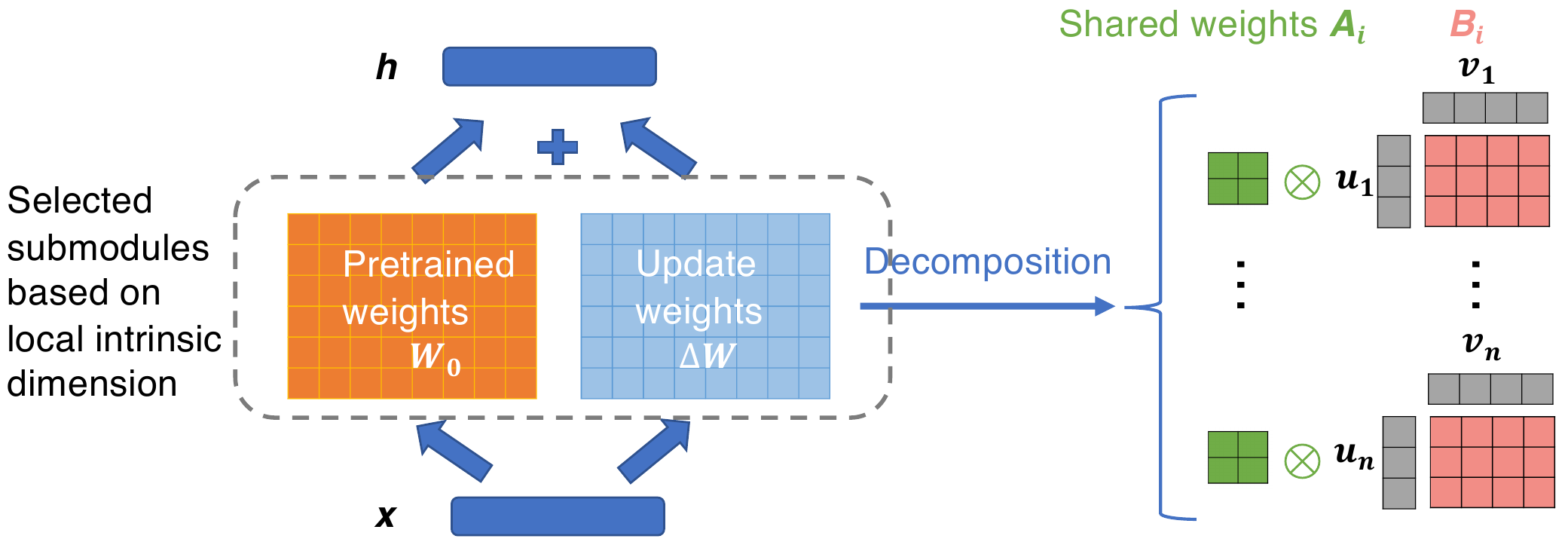}
 	\caption{ An illustration of KAdaptation.  $\boldsymbol{A_i}$ denotes the shared weight matrix, with $i \in\{1, \ldots, n\}$. $\boldsymbol{B_i}$ is decomposed into two low-rank matrices $\boldsymbol{u}_{\boldsymbol{i}}$ and $\boldsymbol{v}_{\boldsymbol{i}}$. $\boldsymbol{h}$ is the output of the selected ViT submodule. $\boldsymbol{x}$ is the input to the submodule. During model adaptation process, only matrices $\boldsymbol{A_i}$, $\boldsymbol{u}_{\boldsymbol{i}}$, and $\boldsymbol{v}_{\boldsymbol{i}}$ receive gradients to improve parameter efficiency.
 	}\label{fig:overview}
 	\vspace{-0.2cm}
	\end{center}
 \end{figure}
\subsubsection{Kronecker Adaptation}
Leveraging the Kronecker product to perform language model compression has been shown to be beneficial in prior works~\cite{kroneckerbert, kroneckergpt}. Recently, Zhang~\etal~\shortcite{phm} introduces PHM layers, theoretically demonstrating that Kronecker products can help to reduce learnable parameters in language models and maintain performance. Built upon the success of PHM, for an update matrix $\Delta\boldsymbol{W} \in \mathbb{R}^{k \times d}$ in the ViT,
we propose the KAdaptation to adapt it into subspace. The illustration is shown in Fig.~\ref{fig:overview}. Mathematically, we compute $\Delta\boldsymbol{W}$ as follows: \begin{equation}
\Delta\boldsymbol{W}=\sum_{i=1}^{n} \boldsymbol{A}_{\boldsymbol{i}} \otimes \boldsymbol{B}_{\boldsymbol{i}},
\label{eq:kronecker}
\end{equation}
where $n$ is the user-defined hyperparameter  representing the number of Kronecker products, $\boldsymbol{A}_{\boldsymbol{i}} \in \mathbb{R}^{n \times n} \text {, and } \boldsymbol{B}_{\boldsymbol{i}} \in \mathbb{R}^{\frac{k}{n} \times \frac{d}{n}}$. The new representation of the update weights in Eq.~\ref{eq:kronecker} is composed of a sum of $n$ Kronecker
products between shared {\em slow weights} $\boldsymbol{A}_i$ and independent {\em fast weights} $\boldsymbol{B}_i$\text {, with } $i \in\{1, \ldots, n\}$.

Meanwhile, low-rank methods~\cite{aghajanyanIntrinsicDimensionalityExplains2020,liMeasuringIntrinsicDimension2018, low_rank} have demonstrated that strong performance can be achieved by optimizing models in a low-rank subspace. Similarly, we hypothesize that $\Delta\boldsymbol{W} $ can be effectively adapted by learning transformations in a low-rank subspace to reduce parameter cost further. Therefore, we parameterize $\boldsymbol{B}_i \in \mathbb{R}^{\frac{k}{n} \times \frac{d}{n}}$ as low rank and further decompose it into the product of two low-rank matrices $\boldsymbol{u}_{\boldsymbol{i}} \in \mathbb{R}^{\frac{k}{n} \times r}$ and $\boldsymbol{v}_{\boldsymbol{i}} \in \mathbb{R}^{r \times \frac{d}{n}}$, where $r$ is the rank of the matrix. Overall, similar to the low-rank parameterized hypercomplex multiplication layer (LPHM) proposed in~\citet{mahabadiCompacterEfficientLowRank2021}, the expression of the update matrix $\Delta\boldsymbol{W} $ is then:
\begin{equation}
    \Delta\boldsymbol{W}=\sum_{i=1}^{n} \boldsymbol{A}_{\boldsymbol{i}} \otimes \boldsymbol{B}_{\boldsymbol{i}}=\sum_{i=1}^{n} \boldsymbol{A}_{\boldsymbol{i}} \otimes\left(\boldsymbol{u}_{\boldsymbol{i}} \boldsymbol{v}_{\boldsymbol{i}}^{\top}\right).
\end{equation}
The number of trainable parameters is now substantially saved. Note that similar to $\boldsymbol{B}_i \in \mathbb{R}^{\frac{k}{n} \times \frac{d}{n}}$, the shared {\em slow weights} $\boldsymbol{A}_i$ can also be further decomposed into the product of low-rank matrices. Additional bias terms can also be applied to the update matrix. We give the analysis of parameter efficiency in the next section.

\section{Analysis of Efficient Model Adaptation Methods} 
\subsection{Discussion of State-of-the-art Methods}
In what follows, we discuss connections between our method and state-of-the-art parameter-efficient tuning methods on NLP tasks and provide additional insight into the characteristics of our method.

\paragraph{Adapter-tuning~\cite{houlsbyParameterEfficientTransferLearning2019}} is  the first efficient model adaptation work in the NLP community. It brings in an additional trainable set of modules by adding a trainable bottleneck layer after the
feedforward network in each Transformer layer of
the pretrained language models. A bottleneck layer consists of a down and up projection pair that shrinks and recovers the size of
token hidden states. 

Similar to the Adapter-tuning method where they use the bottleneck structure in the additional layer, our method implements low-rank decomposition on the {\em fast} rank-one matrices~\cite{fast_weights}. The critical functional difference is that our learned weights can be merged with the main weights during inference, thus introducing no latency.

\paragraph{LoRA~\cite{huLoRALowRankAdaptation2021}} is another line of work for parameter-efficient language model tuning: it treats the model parameters after fine-tuning as an addition of the pretrained parameters $\Theta_{\text {pretrained }}$ and task-specific differences $\theta_{\text {task }}$, where $\Theta_{\text {pretrained }}$ is fixed and a new subset of model parameters are added on top. 
Given a pretrained weight matrix $\boldsymbol{W}_{0} \in \mathbb{R}^{d \times k}$, they constrain its update by 
performing low-rank decomposition: $\boldsymbol{W_{0}}+\Delta \boldsymbol{W}=\boldsymbol{W}_{0}+\boldsymbol{B A}$, where $\boldsymbol{A} \in \mathbb{R}^{r \times k}$, $\boldsymbol{B} \in \mathbb{R}^{d \times r}$, and the rank $r \ll \min (d, k)$. By doing this, the weight matrices are split into two parts, where during training, $\boldsymbol{W}_{0}$ is frozen and receives no gradient updates, while only $\boldsymbol{A}$ and $\boldsymbol{B}$ contain trainable parameters.

Our work differs from LoRA mainly in that we decompose weight updates to a set of Kronecker product decomposition. The decomposed {\em slow weight} are shared across layers, further reducing the parameter cost.

\paragraph{Compacter~\cite{mahabadiCompacterEfficientLowRank2021}} 
inserts task-specific weight matrices into weights of pretrained models. Each Compacter weight matrix is computed as the sum of Kronecker products between shared {\em slow} weights and {\em fast} matrices defined per Compacter layer. 

In a similar vein to Compacter, we also leverage the Kronecker product in our method to reduce parameter cost further. Yet, apart from application domains, our method fundamentally differs from Adapter/Compacter based methods in that: first, our method brings in no additional layer and introduces no latency; Second,
our method first selects submodules by measuring the local intrinsic dimension and then performs the KAdaptation over the update weights to selected submodules; Third, during adaptation, only updates to the weights of selected submodules receive gradients and tuned, while pretrained weights are always fixed.

\begin{table}[t]
  \centering
   \resizebox{0.9\columnwidth}{!}{
  \begin{tabular}{lcc}
    \toprule
    Method & \#Params & Complexity\\
    \midrule
      Adapter-tuning & $4Lkd$&$\mathcal{O}\left(kd\right)$
      \\ LoRA & $2Lrd_{model}$&$\mathcal{O}\left(rd_{model}\right)$
      \\ Compacter &$4L(\frac{k}{n}+\frac{d}{n})+n^3$&$\mathcal{O}\left(\frac{k+d}{n}\right)$
      \\ KAdaptation & $2L(\frac{d_{model}}{n}+\frac{r}{n})+n^3$&$\mathcal{O}\left(\frac{r+d_{model}}{n}\right)$ \\
    \bottomrule
  \end{tabular}}
   \caption{Parameter count in Adapter-tuning, LoRA, Compacter, and KAdaptation. $L$ is the number of layers in the Transformer. $k$ is the size of the input dimension to the Adapter layer. $d$ is the bottleneck dimension in the Adapter layer. $d_{model}$ is the Transformer hidden size. $r$ denotes the rank in the low-rank decomposition step. $n$ is the number of Kronecker products usually very small.}
 \label{tab:num_par}
\end{table}

\subsection{Analysis of Parameter Efficiency} 
We analyze the parameter-efficiency of our KAdaptation and other model adaptation methods as below:

\paragraph{Adapter-tuning} In the standard setting, two Adapters are added per layer of a Transformer model~\cite{Baevski2019AdaptiveIR}. Each Adapter layer consists of $2\times k\times d$ parameters for the down and up-projection matrices, where $k$ is the size of the input dimension and $d$ is the Adapter's bottleneck dimension.
The total number of parameters for Adapters for a $L-$layer Transformer is, $|\Theta| = 2\times L \times 2\times k\times d $s.

\paragraph{LoRA} LoRA adds trainable pairs of rank decomposition matrices to existing weight matrices. The number of trainable parameters is determined by the rank $r$: $|\Theta|=2 \times L \times d_{\text {model }} \times r$, where $d_{\text {model }}$ is Transformer hidden size.
\paragraph{Compacter} Compacter shares the trained weight matrices $\left\{\boldsymbol{A}_{\boldsymbol{i}}\right\}_{i=1}^{n}$ consisting
of $n^{3}$ parameters across all layers, where $n$ is the number of Kronecker products. Compacter also has two rank-one weights for each Adapter layer consisting of $\frac{k}{n}+\frac{d}{n}$ parameters, where the Adapter layers are of size $k \times d$, resulting in a total of $2\times \left(\frac{k}{n}+\frac{d}{n}\right)$ parameters for down and up-projection weights. Therefore, the total number of parameters of Compacter is $4 \times L \times \left(\frac{k}{n}+\frac{d}{n}\right)+n^{3}$ for a Transformer with $L$ layers in the encoder and decoder.

\paragraph{Our Approach} we analyze the parameter efficiency of our approach under the scenario where we decompose the updates to weights into a sum of Kronecker products first and then further perform low-rank decomposition for the {\em fast weights}. The total number of parameters in this scenario will be:
$2 \times L \times \left(\frac{r+d_{model}}{n}\right)+n^{3}$.

The overall comparison of parameter counts is shown in Table~\ref{tab:num_par}. Our method has a complexity of $\mathcal{O}\left(\frac{r+d_{model}}{n}\right)$ with $r$ being a small integer. Our approach greatly reduces the number of parameters. The exact numbers of trainable parameters are present in Table~\ref{tab:acc}.

\section{Experiments}
\begin{table*}[t]
  \centering
  \resizebox{2.12\columnwidth}{!}{
\begin{tabular}{lcccccccccccccccccccccccc}
\toprule
Method &
  \rotatebox[origin=c]{90}{Caltech101} & \rotatebox[origin=c]{90}{CIFAR10} & \rotatebox[origin=c]{90}{CIFAR100} & \rotatebox[origin=c]{90}{Country211} & \rotatebox[origin=c]{90}{DTD} & \rotatebox[origin=c]{90}{EuroSat} & \rotatebox[origin=c]{90}{FER2013} & \rotatebox[origin=c]{90}{FGVCAircraft} & \rotatebox[origin=c]{90}{Food101} & \rotatebox[origin=c]{90}{GTSRB} & \rotatebox[origin=c]{90}{HatefulMemes} & \rotatebox[origin=c]{90}{KittiDistance} & \rotatebox[origin=c]{90}{MNIST} & \rotatebox[origin=c]{90}{Flowers102} & \rotatebox[origin=c]{90}{OxfordPets} & \rotatebox[origin=c]{90}{PatchCamelyon} & \rotatebox[origin=c]{90}{SST2} & \rotatebox[origin=c]{90}{RESISC45} & \rotatebox[origin=c]{90}{StanfordCars} & \rotatebox[origin=c]{90}{VOC2007} &
 \rotatebox[origin=c]{90}{Ave Acc ($\uparrow$)} &
\rotatebox[origin=c]{90}{\#Params ($\downarrow$)} &
\rotatebox[origin=c]{90}{PE ($\uparrow$)} 
  \\ \midrule
 Fine-tuning &
  87.64 &
  \multicolumn{1}{r}{91.11} &
  \multicolumn{1}{r}{71.52} &
  \multicolumn{1}{r}{15.75} &
  \multicolumn{1}{r}{54.36} &
  \multicolumn{1}{r}{85.24} &
  \multicolumn{1}{r}{52.72} &
  \multicolumn{1}{r}{26.22} &
  \multicolumn{1}{r}{83.28} &
  \multicolumn{1}{r}{74.05} &
  \multicolumn{1}{r}{55.64} &
  \multicolumn{1}{r}{39.15} &
  \multicolumn{1}{r}{65.55} &
  \multicolumn{1}{r}{80.55} &
  \multicolumn{1}{r}{87.31} &
  \multicolumn{1}{r}{64.92} &
  \multicolumn{1}{r}{59.09} &
  \multicolumn{1}{r}{75.61} &
  \multicolumn{1}{r}{57.21} &
  \multicolumn{1}{r}{82.95} &
  \multicolumn{1}{r}{65.49} &
  87,878,739 &
  0.498 
  \\ 
Linear-probing &
  90.96 &
  \multicolumn{1}{r}{90.35} &
  \multicolumn{1}{r}{67.31} &
  \multicolumn{1}{r}{17.36} &
  \multicolumn{1}{r}{62.04} &
  \multicolumn{1}{r}{72.95} &
  \multicolumn{1}{r}{51.91} &
  \multicolumn{1}{r}{29.52} &
  \multicolumn{1}{r}{83.82} &
  \multicolumn{1}{r}{56.47} &
  \multicolumn{1}{r}{55.83} &
  \multicolumn{1}{r}{40.37} &
  \multicolumn{1}{r}{77.50} &
  \multicolumn{1}{r}{92.29} &
  \multicolumn{1}{r}{88.03} &
  \multicolumn{1}{r}{59.00} &
  \multicolumn{1}{r}{59.36} &
  \multicolumn{1}{r}{78.10} &
  \multicolumn{1}{r}{68.30} &
  \multicolumn{1}{r}{84.99} &
  \multicolumn{1}{r}{\textcolor{dblue}{66.32}} &
  \textbf{29,523} &
  \textcolor{dblue}{0.663} 
  \\
Adapter-tuning &
  90.18 &
  \multicolumn{1}{r}{90.14} &
  \multicolumn{1}{r}{73.57} &
  \multicolumn{1}{r}{16.83} &
  \multicolumn{1}{r}{57.13} &
  \multicolumn{1}{r}{67.97} &
  \multicolumn{1}{r}{41.76} &
  \multicolumn{1}{r}{30.52} &
  \multicolumn{1}{r}{83.58} &
  \multicolumn{1}{r}{58.50} &
  \multicolumn{1}{r}{48.91} &
  \multicolumn{1}{r}{37.18} &
  \multicolumn{1}{r}{80.34} &
  \multicolumn{1}{r}{90.78} &
  \multicolumn{1}{r}{86.52} &
  \multicolumn{1}{r}{59.92} &
  \multicolumn{1}{r}{58.70} &
  \multicolumn{1}{r}{79.22} &
  \multicolumn{1}{r}{67.68} &
  \multicolumn{1}{r}{82.22} &
  \multicolumn{1}{r}{65.08} &
 1,237,587  &
  0.647 
  \\
LoRA &
  87.64 &
  \multicolumn{1}{r}{90.52} &
  \multicolumn{1}{r}{69.69} &
  \multicolumn{1}{r}{17.12} &
  \multicolumn{1}{r}{50.16} &
  \multicolumn{1}{r}{74.03} &
  \multicolumn{1}{r}{51.04} &
  \multicolumn{1}{r}{20.01} &
  \multicolumn{1}{r}{83.76} &
  \multicolumn{1}{r}{42.96} &
  \multicolumn{1}{r}{55.88} &
  \multicolumn{1}{r}{48.05} &
  \multicolumn{1}{r}{61.36} &
  \multicolumn{1}{r}{74.28} &
  \multicolumn{1}{r}{85.49} &
  \multicolumn{1}{r}{63.20} &
  \multicolumn{1}{r}{57.04} &
  \multicolumn{1}{r}{62.09} &
  \multicolumn{1}{r}{54.89} &
  \multicolumn{1}{r}{80.33} &
  \multicolumn{1}{r}{61.48} &
 176,979 &
  0.614 
  \\
Compacter &
  89.02 &
  \multicolumn{1}{r}{79.96} &
  \multicolumn{1}{r}{44.33} &
  \multicolumn{1}{r}{28.22} &
  \multicolumn{1}{r}{52.93} &
  \multicolumn{1}{r}{50.48} &
  \multicolumn{1}{r}{35.46} &
  \multicolumn{1}{r}{41.13} &
  \multicolumn{1}{r}{78.28} &
  \multicolumn{1}{r}{66.90} &
  \multicolumn{1}{r}{47.60} &
  \multicolumn{1}{r}{57.72} &
  \multicolumn{1}{r}{85.82} &
  \multicolumn{1}{r}{88.29} &
  \multicolumn{1}{r}{79.23} &
  \multicolumn{1}{r}{61.83} &
  \multicolumn{1}{r}{64.22} &
  \multicolumn{1}{r}{63.76} &
  \multicolumn{1}{r}{64.79} &
  \multicolumn{1}{r}{75.84} &
  \multicolumn{1}{r}{62.79} &
 77,907&
  0.628 \\
  \hdashline
  KAdaptation &
  { 88.96} &
 \multicolumn{1}{r}{{ 90.03}} &
  \multicolumn{1}{r}{{ 73.92}} &
  \multicolumn{1}{r}{{ 17.53}} &
  \multicolumn{1}{r}{{ 63.97}} &
  \multicolumn{1}{r}{{ 76.25}} &
  \multicolumn{1}{r}{{ 47.45}} &
  \multicolumn{1}{r}{{ 30.04}} &
  \multicolumn{1}{r}{{ 84.38}} &
  \multicolumn{1}{r}{{ 80.71}} &
  \multicolumn{1}{r}{{ 55.86}} &
  \multicolumn{1}{r}{{ 42.29}} &
  \multicolumn{1}{r}{{ 85.20}} &
  \multicolumn{1}{r}{{ 93.19}} &
  \multicolumn{1}{r}{{ 89.05}} &
  \multicolumn{1}{r}{{ 63.39}} &
  \multicolumn{1}{r}{{ 59.18}} &
  \multicolumn{1}{r}{{ 79.96}} &
  \multicolumn{1}{r}{{ 70.21}} &
  \multicolumn{1}{r}{{ 84.49}} &
  \multicolumn{1}{r}{{ \textbf{68.92}}} &
  \textcolor{dblue}{79,699}&
  \textbf{0.689}
  \\
  \bottomrule
\end{tabular}
}
 \caption{ The averaged 5-shot experimental result comparison on 20 datasets from ELEVATER benchmark~\cite{vision_benchmark} in terms of accuracy (\%) and number of trainable parameters (\#Params) across random seeds of \{$0$, $1$, $2$\}. The vision transformer (ViT-B-224/32) via CLIP pretraining is evaluated.  Our method achieves the best tradeoff between accuracy and parameter efficiency: it obtains the best average accuracy among all efficient model adaptation methods, while updating only 0.09\% of the model parameters in CLIP. We color each accuracy value as the \textbf{best} and \textcolor{dblue}{second best}: the same hereinafter.} 
 \label{tab:fewshot}
\end{table*}
\subsection{Datasets} 
For few-shot benchmark experiments, we conduct experiments on 20 image classification datasets from the ELEVATER benchmark~\cite{vision_benchmark} on four Quadro RTX A6000 GPUs. Detailed dataset statistics are given in the supplementary material. For full-shot experiments, we summarize the results by computing the average performance on CIFAR10~\cite{cifar}, CIFAR100~\cite{cifar}, SUN397~\cite{sun}, DTD~\cite{dtd}, STL10~\cite{stl10}, FGVCAircraft~\cite{fgvc}, and FER2013~\cite{fer}. We use the official split for each of these datasets.

\subsection{Implementation Details}
For benchmark experiments, we use the SGD~\cite{sgd} optimizer with the learning rate and weight decay being automatically searched for all methods so that these two hyperparameters have the optimum combination. We borrow the automatic hyper-parameter tuning toolkit from~\citet{vision_benchmark}. Training epochs are set via grid search. We test two pretrained $12$-layer ViTs: the one using ViT-B-224/32 via unsupervised pretraining ({\em CLIP}) and the one using ViT-B-224/16 via supervised pretraining ({\em Supervised ViT}).

For intrinsic dimension experiments, we use the AdamW~\cite{adam} as the optimizer, with the weight decay of $10^{-8}$, learning rate of $10^{-5}$, and batch size of 32 following the setting in Li~\etal~\shortcite{liMeasuringIntrinsicDimension2018}. The Fastfood transform~\cite{le2014fastfood} is applied to the attention and multi-layer perceptron (MLP) module in the first layer of Supervised ViT, respectively. The dimension $d$ is measured from $0-2000$ in both scenarios. Each model is fine-tuned for 300 epochs.

\subsection{Baselines}
We test the baselines below. Unless otherwise specified, the task-specific classification layer and added parameters are tuned while the pretrained ViTs are frozen.

First are commonly-used model adaptation methods for vision models.
\begin{itemize}
\item \emph{Full-model Fine-tuning}: fine-tunes all model parameters.
\item \emph{Linear-probing}: only tune the task-specific classification layer. 
\end{itemize}

The second types are SOTA methods borrowed from the NLP community.

\begin{itemize}
\item \emph{BitFit}~\cite{zakenBitFitSimpleParameterefficient2021}: freezes all ViT parameters except for the bias terms and the task-specific classification layer.
\item \emph{Adapter}-tuning~\cite{houlsbyParameterEfficientTransferLearning2019}: two Adapters are added and tuned in each Transformer layer.
\item \emph{AdapterDrop}~\cite{ruckleAdapterDropEfficiencyAdapters2021}: only keep Adapters from the last Transformer layer.
\item \emph{LoRA}~\cite{huLoRALowRankAdaptation2021}: apply LoRA to $\boldsymbol{W}_q$ and $\boldsymbol{W}_v$ matrices in the attention module and tune the low-rank decomposition matrices.
\item \emph{Compacter}~\cite{mahabadiCompacterEfficientLowRank2021}: we experiment with $n=4$.
\end{itemize}

The third types are new baseline methods we developed.

\begin{itemize}
\item \emph{Transformer-probing}: an additional trainable Transformer block is stacked before the task-specific classification layer and tuned.
\item \emph{LoRA-Fix}: the matrix $\boldsymbol{A}$ in LoRA~\cite{huLoRALowRankAdaptation2021} is fixed and only the matrix $\boldsymbol{B}$ is tuned.
\item \emph{LayerNorm Tuning}: the layer norm layers are tuned.
\item\emph{Attention Tuning}: the attention layers are tuned.
\item \emph{LePE Tuning}~\cite{dong2021cswin}: locally-enhanced positional encoding (LePE) is added to the ViT and tuned. We implement it by the depthwise convolution operator~\cite{depthwidth} on the matrix $\boldsymbol{V}$ in the attention layer:
$
\operatorname{Attention}(\boldsymbol{Q}, \boldsymbol{K}, \boldsymbol{V})= \operatorname{SoftMax}\left(\boldsymbol{Q} \boldsymbol{K}^{T} / \sqrt{d}\right) \boldsymbol{V}+\operatorname{DWConv}(\boldsymbol{V}).$
\item \emph{Relative Position Bias (RPB) Tuning}~\cite{liuSwinTransformerHierarchical2021}: an additional relative position bias term $\boldsymbol{B}$ is included in computing self-attention in the ViT and tuned:
$
\operatorname{Attention}(\boldsymbol{Q}, \boldsymbol{K}, \boldsymbol{V})= \operatorname{SoftMax}\left(\boldsymbol{Q} \boldsymbol{K}^{T} / \sqrt{d}+\boldsymbol{B}\right)\boldsymbol{V}.
$
\end{itemize}
LayerNorm Tuning, Attention Tuning, and BitFit shed light on which parameters in ViT matter more during model adaptation. Among all modules in ViT, multi-layer perceptron (MLP) tuning is not considered a baseline because it is prohibitively costly compared to others. Given that the special structure of ViT and its variants, e.g., depthwise convolution operator and relative position bias, are different from the general transformers in natural language processing,  we actually made the first step towards parameter-efficient model adaptation for the ViT
via LePE Tuning and Relative Position Bias Tuning.

\subsection{Results and Analysis}
\paragraph{Metric with performance-efficiency trade-off}
To better compare different methods with a single number that considers both prediction accuracy and parameter-efficiency, we resort to the performance-efficiency (PE) metric defined in~\citet{li2022elevater}:
$$
\text {PE}=\text {score} * \exp \left(-\log _{10}(\text {\# trainable-parameters } / M_0+1)\right)
$$ where score is the prediction accuracy, while \# trainable-parameters is the number of updated parameters in the model adaptation stage, and $M_0$ is the normalization constant. $M_0$ is set to $10^8$ because most existing vision backbone model size are in this magnitude, for example, ViT-Base (80M parameters). 

The experimental results of measured average accuracy across the 20 datasets in the low-data regime and under the 5-shot setting using random seeds of $0$, $1$, and $2$ are shown in Table~\ref{tab:fewshot}. As observed, the parameter cost of linear-probing is the lowest while that of full-model fine-tuning is the highest. Our method has the highest average accuracy and remains the ideal approach with the optimum tradeoff: our method has much less trainable parameters than other adaptation methods --- the second lowest and is only higher than Linear-probing. From the performance-efficiency trade-off metric, it can also be seen that \textbf{ours has the highest PE}.

To further compare our method with SOTA methods for NLP models and more baselines, we investigate the performance of adaptation approaches in the full-data regime and test under the full-shot setting. The results across the seven datasets are shown in Table~\ref{tab:acc}. In our analytical experiments, we first observe that Full-model Fine-tuning has the highest accuracy in both scenarios, serving as a performance upper bound. Second, different efficient model adaptation methods exhibit diverse characteristics
and perform differently on the same task. Third, the results from CLIP are mostly consistent with the results from Supervised ViT. This suggests that the pretraining strategy may not affect the selection of downstream model adaptation strategy much. Fourth, previous methods such as Adapter-tuning~\cite{houlsbyParameterEfficientTransferLearning2019} and LoRA~\cite{huLoRALowRankAdaptation2021} are still effective, and their accuracy is substantially higher than naive baselines, including BitFit and Attention-tuning regardless of the pretrained checkpoint. 
Fifth, among naive baselines where only submodules or task-specific classification heads are tuned, tuning the parameters of the attention layer turns out to be a surprisingly effective approach even compared to some SOTA methods, though its parameter cost is significantly higher. This further validates the effectiveness of our method by applying KAdaptation to attention weights. Finally, our method outperforms all the SOTA methods borrowed from the NLP community as well as their variants in both scenarios.

Furthermore, the average number of trainable parameters across seven datasets is also shown in Table~\ref{tab:acc}. As can be seen, our KAdaptation method contains the lowest parameter cost compared with other SOTA methods. This phenomenon is obviously noticeable when compared with Full-model Fine-tuning, where our method takes less than 0.14\% of trainable parameters of end-to-end Full-model Fine-tuning but it is capable of achieving comparable performance.

\begin{table*}[t]
\small
  \resizebox{2.12\columnwidth}{!}{
  \centering
  \setlength{\tabcolsep}{2.5pt}
  \begin{tabular}{l ccccccccc  ccccccccc}
    \toprule
    \multirow{2}*{Method} & \multicolumn{9}{c}{CLIP} & \multicolumn{9}{c}{Supervised ViT}\\
    \cmidrule(lr){2-10} \cmidrule(lr){11-19}&CIFAR10&CIFAR100&SUN397&DTD&FER2013&FGVCAircraft&STL10&Average ($\uparrow$) &\#Params ($\downarrow$) &CIFAR10&CIFAR100&SUN397&DTD&FER2013&FGVCAircraft&STL10&Average ($\uparrow$)&\#Params ($\downarrow$)\\
    \midrule
     \multicolumn{19}{c}{Commonly-used model adaptation methods for vision models} \\
    \midrule
   Full-model Fine-tuning &\textbf{97.7}&\textbf{85.4}&73.8&\textbf{79.0}&\textbf{69.8}&\textbf{59.0}&\textbf{99.7}&\textbf{80.6} &87,897,654 &\textbf{99.0}&\textbf{92.4}&75.0&72.4&\textbf{68.2}&52.6&\textbf{99.6}&\textbf{79.9} & 86,630,561\\
      Linear-probing & 94.8&80.1&72.4&75.4&67.3&49.7&98.4&76.9 &49,175 &96.3&87.7&70.1&\textbf{72.7}&60.1&45.0&98.7&75.8 &49,175
      \\
      \midrule
      \multicolumn{19}{c}{SOTA methods for NLP models} \\
      \midrule
      BitFit&92.1&76.0&70.8 &75.9&68.0&54.5&98.8&76.6 & 179,049 &92.3&81.0&71.8&\textcolor{dblue}{72.6}&60.4&45.9&99.0&74.7 &358,741
      \\Adapter-tuning&94.7&81.4&\textcolor{dblue}{77.1} &78.0&68.4&55.3&99.0&79.1 &1,242,843 &98.4&90.6&74.2&71.0&63.4&52.4&99.3&78.5 &1,505,654
      \\AdapterDrop &93.3&78.3&71.4&77.1&67.1&51.3&98.0&76.6 &91,487 &96.8&88.4&72.3&70.2&46.9&35.6&99.6&72.8 &174,646
      \\ LoRA &95.1&78.1&\textbf{80.8}&78.1&67.7 &55.8&99.2&79.3 &147,236 &\textcolor{dblue}{98.7}&90.6&73.6&70.4&62.7&\textcolor{dblue}{54.9}&99.4&78.6 & 219,601\\
      \midrule
      \multicolumn{19}{c}{Baseline methods developed in this work} \\
   \midrule
      Transformer-probing &95.6&80.1&74.3&75.9&67.6&50.9&98.5&77.6 &3,198,999&96.5&86.9&\textbf{76.7}&72.0&60.7&45.5&99.0&76.8 &3,198,999
      \\ LoRA-Fix&92.5&77.1&60.0&77.7&65.5&44.4&88.6&72.3 & 98,481 &96.2&88.3&72.0&65.5&53.4&51.7&99.0&75.2 & 148,704
      \\ LayerNorm Tuning &82.5&76.6&66.7&72.4&61.0&37.6&99.1&70.8& 52,405 &92.2&71.7&72.0&69.0&52.7&51.0&98.8&72.5 &75,413
      \\  Attention Tuning &\textcolor{dblue}{96.8}&81.8&73.1&75.0&62.2&54.2&97.6&77.2&41,005,636 &93.9&85.7&73.8&69.2&55.2&51.9&99.2&75.6 &28,405,278
      \\
      LePE Tuning &95.1&78.9&68.0&75.4&65.2&54.0&98.0&76.4&112,556
      &93.7&90.8&73.2&69.8&60.0&49.3&99.1& 76.6 &167,225
      \\RPB Tuning &94.7&77.1& 68.4&75.2&65.1&54.1&97.9&76.1& 66,768
      &96.7&87.0&72.4&70.4&50.9&51.4&98.9&75.4&145,920
      \\ 
       \midrule
       KAdaptation &95.9&\textcolor{dblue}{84.8}&74.0&\textcolor{dblue}{78.1}&\textcolor{dblue}{69.0}&\textcolor{dblue}{56.0}&\textcolor{dblue}{99.2}&\textcolor{dblue}{79.6} &80,726 &97.9&\textcolor{dblue}{91.2}&\textcolor{dblue}{75.1}&71.4&\textcolor{dblue}{63.8}&\textbf{55.5}&\textcolor{dblue}{99.4}&\textcolor{dblue}{79.2} & 114,079\\
    \bottomrule
  \end{tabular}
  }
   \caption{Experimental result comparison on CIFAR10~\cite{cifar}, CIFAR100~\cite{cifar}, SUN397~\cite{sun}, DTD~\cite{dtd}, STL10~\cite{stl10}, FGVCAircraft~\cite{fgvc}, and FER2013~\cite{fer} datasets in terms of accuracy (\%) and number of trainable parameters (\#Params). 
 }
 \label{tab:acc}
\end{table*}

To further validate the efficiency of our proposed method, in addition to parameter costs, we perform additional evaluation on memory footprint and inference time. We compare the per-sample memory usage of each method in Table~\ref{tab:time}. Our method reduces memory overhead by $-86.0$\% compared to Full-model Fine-tuning and is in the same order of magnitude as other efficient model adaptation methods. We compare the inference time cost per batch in Table~\ref{tab:time} as well. On average,
our method costs $6.93$s per batch, the same as the vanilla ViT and LoRA, while Adapter-tuning costs $12.97$s and
Compacter takes $14.90$s.  Our method is the most efficient. It's within expectation as our method does not bring any additional layer to the original ViT, suffering from no inference latency. 

    \begin{table}[t]
\centering
  \resizebox{ \columnwidth}{!}{
         \begin{tabular}{lccc}
    \toprule
    Method   & Average Accuracy ($\uparrow$)  & Inference time ($\downarrow$)   & Memory ($\downarrow$)    \\
    \midrule
     Full-model Fine-tuning&\textbf{79.9} &6.93&421.5  
     \\ Linear-probing&   75.8 & 6.93 &\textbf{27.1}  
  \\  Adapter-tuning&  78.5 & \textcolor{dblue}{12.97}& 70.2 
      \\ LoRA& 78.6 &6.93&\textcolor{dblue}{56.0} 
  \\ Compacter&  78.6 & 14.90  & 70.0\\
   \hdashline
       KAdaptation&\textcolor{dblue}{79.2}&\textbf{6.93} &59.1  \\
    \bottomrule
\end{tabular}}
\caption{Average accuracy (\%), average inference time/throughput (s) per batch, and average peak memory (MB) for each method. Our method is time-efficient, and our memory footprint is in the same order of magnitude as other efficient model adaptation methods and much less than Full-model Fine-tuning.}
\label{tab:time}
\end{table}

\subsection{Local Intrinsic Dimension}
\begin{figure}[t]
	\begin{center}
 	\includegraphics[width = 0.7\columnwidth]{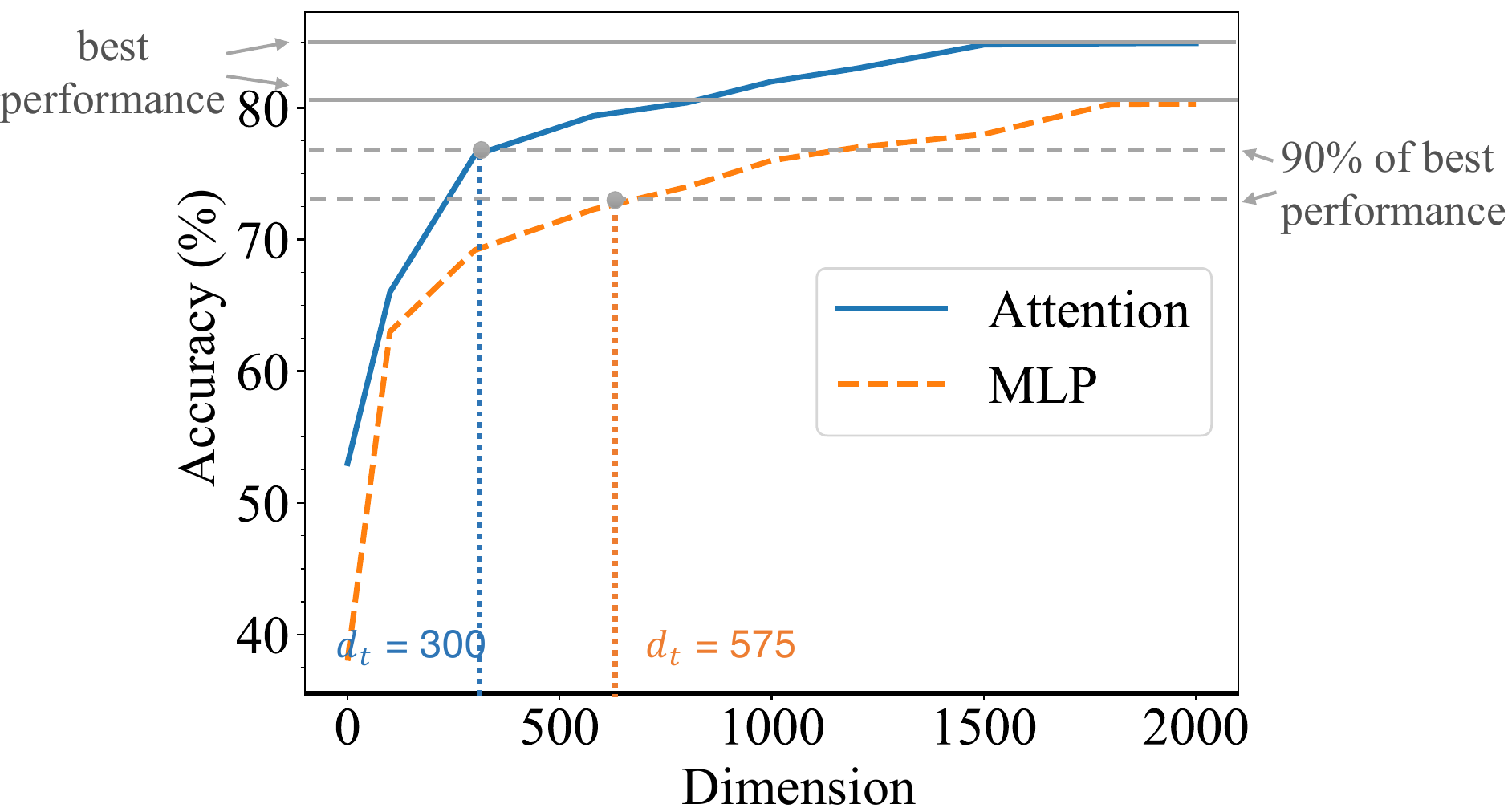}
 	\caption{Validation Accuracy \textit{vs.} Subspace Dimension $d$ of MLP and the attention module for Supervised ViT on CIFAR100. The local intrinsic dimension $d_t$ of the attention module is lower than that of the MLP.}\label{fig:dim}
	\vspace{-0.2cm}
	\end{center}
 \end{figure}

Local intrinsic dimension~\cite{liMeasuringIntrinsicDimension2018} informs us of the importance of each module in the ViT and we select submodules to perform KAdaptation based on the measurement results of the local intrinsic dimension. We measure the local intrinsic dimension of the two fundamental architectural components in the ViT --- the MLP module and the attention module. We use the remarkable Fastfood transform~\cite{le2014fastfood} to do the projection. The accuracy results averaged across $\{1, 6, 12\}$-th ViT layers are shown in Fig.~\ref{fig:dim}. As a substantiating point to performing Kronecker Adaptation on attention layers, we can see the attention module has a lower intrinsic dimension than the MLP module (300 \textit{vs.} 575).

 \begin{table}[t]
\centering
  \resizebox{0.7\columnwidth}{!}{
  \begin{tabular}{lc}
    \toprule
    Method & \ Average Accuracy \\
    \midrule
      Adapters on attention layer & 54.1 \\
      Standard Adapter-tuning & 87.7 \\
      KAdaptation to MLP & 86.6\\
    KAdaptation  & 88.1\\
    \bottomrule
  \end{tabular}}
  \caption{KAdaptation and Adapter-tuning ablation experiments with Supervised ViT on CIFAR10~\cite{cifar}, CIFAR100~\cite{cifar}, and SUN397~\cite{sun}. We report the average accuracy (\%) across the three datasets.}
  \label{tab:ablation1}
  \vspace{-0.2cm}
  \end{table}

 \subsection{Ablation Studies}
 We ablate our method and Adapter-tuning using the settings in Table~\ref{tab:acc}. As can be seen in Table~\ref{tab:ablation1}, several intriguing properties are observed.
First, applying KAdaptation to MLP modules performs worse than the original method where we apply KAdaptation to attention modules. This phenomenon is consistent with our findings from naive baseline experiments and intrinsic dimension experiments.
Second, we test another variant of Adapter-tuning. Instead of inserting two Adapters after the attention and feedforward modules respectively following Houlsby~\etal~\shortcite{houlsbyParameterEfficientTransferLearning2019}, we add Adapters in the attention layers. It can be observed that the standard Adapter-tuning outperforms this variance, indicating the effectiveness of the vanilla Adapter-tuning when it is adapted to vision tasks.

\section{Conclusion}
In this paper, we conduct the first comprehensive comparison of efficient model adaptation on the image classification tasks using vision transformers. We also propose a better parameter-efficient model adaptation strategy in the principle of subspace training and parameterized hypercomplex multiplication, which achieves the best tradeoff between accuracy and parameter efficiency. We release a benchmark by providing the implementation of all the methods studied in this paper, which could be directly used in developing future efficient model adaptation strategies and will hopefully facilitate research in this area. Looking into the future, we plan to explore the generalization of our method to other tasks, especially in the vision-and-language domain.

\bibliography{aaai23}



\bigskip

\appendix

\section{Detailed Dataset Statistics}
For the few-shot setting, we test on  Caltech101~\cite{fei2004learning}, CIFAR10~\cite{cifar}, CIFAR100~\cite{cifar}, Country211~\cite{radford2021learning}, DTD~\cite{cimpoi2014describing}, EuroSat~\cite{helber2019eurosat}, FER2013~\cite{fer2013},  FGVCAircraft~\cite{maji2013fine}, Food101~\cite{bossard2014food}, GTSRB~\cite{stallkamp2011german}, HatefulMemes~\cite{kiela2020hateful}, KittiDistanc~\cite{fritsch2013new},  MNIST~\cite{deng2012mnist}, Flowers102~\cite{nilsback2008automated}, OxfordPets~\cite{parkhi2012cats}, PatchCamelyon~\cite{veeling2018rotation}, SST2~\cite{radford2021learning}, RESISC45~\cite{cheng2017remote}, StanfordCars~\cite{krause20133d}, and VOC2007~\cite{everingham2010pascal} using the vision transformer ({\em ViT-B-224/32}) via Contrastive Language-Image
Pretraining (also known as CLIP). In Table~\ref{table:downstream_ic_dataset}, we list the basic statistics of 20 image classification datasets used in our few-shot setting experiments.

For the full-shot setting, we use the seven datasets mentioned in the main paper: CIFAR10~\cite{cifar}, CIFAR100~\cite{cifar}, SUN397~\cite{sun}, DTD~\cite{dtd}, STL10~\cite{stl10}, FGVCAircraft~\cite{fgvc}, and FER2013~\cite{fer}.

\begin{table*}[h!]
  \centering
\begin{tabular}{c | c c c c c  } 
 \toprule
 Dataset & \#Concepts & Train size & Test size & Evaluation metric   \\ 
 \midrule

Hateful Memes~\cite{kiela2020hateful} & 2 & 8,500 & 500 & ROC AUC  \\
PatchCamelyon~\cite{veeling2018rotation}  & 2 & 262,144 & 32,768 & Accuracy  \\
Rendered-SST2~\cite{radford2021learning} & 2 & 6,920 & 1,821 & Accuracy \\
KITTI Distance~\cite{fritsch2013new} & 4 & 6,347 & 711 & Accuracy  \\
FER 2013~\cite{fer2013} & 7 & 28,709 & 3,589 & Accuracy  \\
CIFAR10~\cite{cifar} & 10 & 50,000 & 10,000 & Accuracy \\
EuroSAT~\cite{helber2019eurosat} & 10 & 5,000 & 5,000 & Accuracy  \\
MNIST~\cite{deng2012mnist} & 10 & 60,000 & 10,000 & Accuracy  \\
VOC 2007 Classification~\cite{everingham2010pascal} & 20 & 2,501 & 4,952 &11-point mAP\\
Oxford-IIIT Pets~\cite{parkhi2012cats} & 37 & 3,680 & 3,669 & Mean-per-class  \\
GTSRB~\cite{stallkamp2011german} & 43 & 26,640 & 12,630 & Accuracy  \\
Resisc-45~\cite{cheng2017remote} & 45 & 3,150 & 25,200 & Accuracy  \\
Describable Textures~\cite{cimpoi2014describing} & 47 & 1,880 & 1,880 & Accuracy  \\
CIFAR100~\cite{cifar} & 100 & 50,000 & 10,000 & Accuracy  \\
FGVC Aircraft (variants)~\cite{maji2013fine} & 100 & 3,334 & 3,333 & Mean-per-class  \\
Food-101~\cite{bossard2014food} & 101 & 75,750 & 25,250 & Accuracy  \\
Caltech101~\cite{fei2004learning} & 102 & 3,060 & 6,084 & Mean-per-class  \\
Oxford Flowers 102~\cite{nilsback2008automated} & 102 & 1,020 & 6,149 & Mean-per-class  \\
Stanford Cars~\cite{krause20133d} & 196 & 8,144 & 8,041 & Accuracy  \\
Country-211~\cite{radford2021learning} & 211 & 31,650 & 21,100 & Accuracy  \\
\midrule
Total & 2151 & 1,919,596 & 242,677 & --  &    \\ 
\bottomrule
\end{tabular}
\caption{Statistics of 21 datasets used in few-shot image classification experiments.}
\label{table:downstream_ic_dataset}
\end{table*}

\begin{table}[tbp]
    \centering
    \resizebox{\columnwidth}{!}{
    \begin{tabular}{lcl}
    \toprule
         Name & Value & Description \\
         \midrule
         Optimizer & SGD &-\\
         Learning rate &$10^{-6}$-$10^6$& Automatically searched for each dataset \\
         Weight decay &-& Automatically searched for each dataset\\
        Max epoch &50&-\\
        Batch size &32&-\\
            \bottomrule
    \end{tabular}}
        \caption{Hyperparameter settings for benchmark experiments under the few-shot setting.}
          \label{tab:Baseline_hyper-paramter_few_shot}
\end{table}

\section{Hyperparameter Settings}
The overall hyperparameter settings for benchmark experiments under the few-shot setting are specified in Table~\ref{tab:Baseline_hyper-paramter_few_shot}. We repeat experiments for three different random seeds of $\{0, 1, 2\}$. The learning rate and weight decay are automatically searched so that these two hyperparameters will always have the optimum combination. The number of seaching epochs is set to 10. The number of training epochs is set to 50. For the method-specific hyperparameter settings, for all Adapter-based methods~\cite{houlsbyParameterEfficientTransferLearning2019}, we experiment with Adapters of bottleneck size of 64; For Compacter~\cite{mahabadiCompacterEfficientLowRank2021}, we experiment with the number of Kronecker products $n=4$; For LoRA~\cite{huLoRALowRankAdaptation2021}, we follow the same setting in the original paper; In our KAdaptation method, we set the user-defined hyperparameter $n \in \mathbb{Z}>0$, which is the number of Kronecker products, by performing the grid search from $\{4, 32, 64\}$ and choosing the best one. We report the model performing the best with $n = 32$.

The overall hyperparameter settings for benchmark experiments under the full-shot setting are specified in Table~\ref{tab:Baseline_hyper-paramter}.  For each method, we set the batch size to 64. All experiments are performed with the same random seed. The optimum number of training epochs is set by grid search in the range of \{100, 200, 400\}.

\begin{table}[tbp]
    \centering
    \resizebox{\columnwidth}{!}{
    \begin{tabular}{lcl}
    \toprule
         Name & Value & Description \\
         \midrule
         Optimizer & SGD &-\\
         Learning rate &$10^{-6}$-$10^6$& Automatically searched for each dataset \\
         Weight decay &-& Automatically searched for each dataset\\
        Max epoch &100-400&Grid search\\
        Batch size &64&-\\
            \bottomrule
    \end{tabular}}
        \caption{Hyperparameter settings for benchmark experiments under the full-shot setting.}
          \label{tab:Baseline_hyper-paramter}
\end{table}

\section{Additional Related Works}
\paragraph{Transformers}
Transformer~\cite{Transformers} is a sequence-to-sequence architecture that
makes heavy use of self-attention mechanisms to replace the recurrence and convolution operations. It is initially used for machine translation~\cite{bahdanau2014neural} and has shown outstanding
performance in a wide range of natural language processing (NLP) tasks, including reading comprehension~\cite{dai2020funnel}, question
answering~\cite{min2019knowledge}, vision-and-language tasks~\cite{li2020oscar}, etc. Since Radford~\etal~\shortcite{gpt2} first applied a stack of Transformer decoders to autoregressive language modeling, fine-tuning Transformer-based language models has dominated NLP, achieving state-of-the-art performance in many tasks. Meanwhile, fine-tuning large-scale pretrained ViTs has shown prominent performance for computer vision tasks, such as image classification~\cite{vision_transformer}, object detection~\cite{detection_transformer}, etc. Recently, there are also other variants, including hierarchical ViTs with varying resolutions and spatial embeddings~\cite{liuSwinTransformerHierarchical2021,dong2021cswin} been proposed. Beyond doubt, the recent progress of large Transformer-alike models posts great demands for developing efficient model adaptation strategies. 

\paragraph{Pretraining}
Pretraining in NLP~\cite{liu2019roberta,bert}, has achieved state-of-the-art performances in many downstream tasks~\cite{thongtan-phienthrakul-2019-sentiment,white2017inference}.
 It is widely deemed that adapting models pretrained on general domain data~\cite{bert,gpt2} to downstream datasets could provide a substantial performance gain compared to training on task-specific data directly. In computer vision, adapting pretrained models, e.g.~\cite{VGG,resnet, vision_transformer}, has come to the forefront of deep learning techniques due to its success in downstream tasks~\cite{ren2015faster,image_text_matching2} that can substantially outperform tuning models with random initialization. In this paper, we will mainly focus on the parameter-efficient model adaptation of pretrained ViT on the image classification task.
\end{document}